\documentclass{article}
\usepackage{ijcai26}

\usepackage{times}
\usepackage{amssymb}
\usepackage{soul}
\usepackage{url}
\usepackage[hidelinks]{hyperref}
\usepackage[utf8]{inputenc}
\usepackage[small]{caption}
\usepackage{graphicx}
\usepackage{amsmath}
\usepackage{amsthm}
\usepackage{booktabs}
\usepackage{algorithm}
\usepackage{algorithmic}
\usepackage[switch]{lineno}
\usepackage{multirow}
\usepackage{graphicx}      
\usepackage[table]{xcolor} 
\usepackage{array}         

\theoremstyle{definition}
\newtheorem{definition}{Definition}

\title{To Know is to Construct: Schema-Constrained Generation for Agent Memory}

\author{
Lei Zheng$^1$
\and
Weinan Song$^1$
\and
Daili Li$^1$
\And
Yanming Yang$^1$\\
\affiliations
$^1$UnionPay\\
\emails
zhenglei2@unionpay.com,
playinlife@126.com,
lidaili@unionpay.com,
ymyang@unionpay.com
}

\begin{document}

\maketitle

\begin{abstract}
Constructivist epistemology argues that knowledge is actively constructed rather than passively copied. 
Despite the generative nature of Large Language Models (LLMs), most existing agent memory systems are still based on dense retrieval. 
However, dense retrieval heavily relies on semantic overlap or entity matching within sentences. 
Consequently, embeddings often fail to distinguish instances that are semantically similar but contextually distinct, introducing substantial noise by retrieving context-mismatched entries. 
Conversely, directly employing open-ended generation for memory access risks "Structural Hallucination"—where the model generates memory keys that do not exist in the memory, leading to lookup failures. 
Inspired by this epistemology, we posit that memory is fundamentally organized by cognitive schemas, and valid recall must be a generative process performed within these schematic structures.
To realize this, we propose SCG-MEM, a schema-constrained generative memory architecture. SCG-MEM reformulates memory access as Schema-Constrained Generation. 
By maintaining a dynamic Cognitive Schema, we strictly constrain LLM decoding to generate only valid memory entry keys, providing a formal guarantee against structural hallucinations. 
To support long-term adaptation, we model memory updates via assimilation (grounding inputs into existing schemas) and accommodation (expanding schemas with novel concepts). 
Furthermore, we construct an Associative Graph to enable multi-hop reasoning through activation propagation. 
Experiments on the LoCoMo benchmark show that SCG-MEM substantially improves performance across all categories over retrieval-based baselines. 

\end{abstract}

\begin{figure*}[t]
    \centering
    \includegraphics[width=\textwidth]{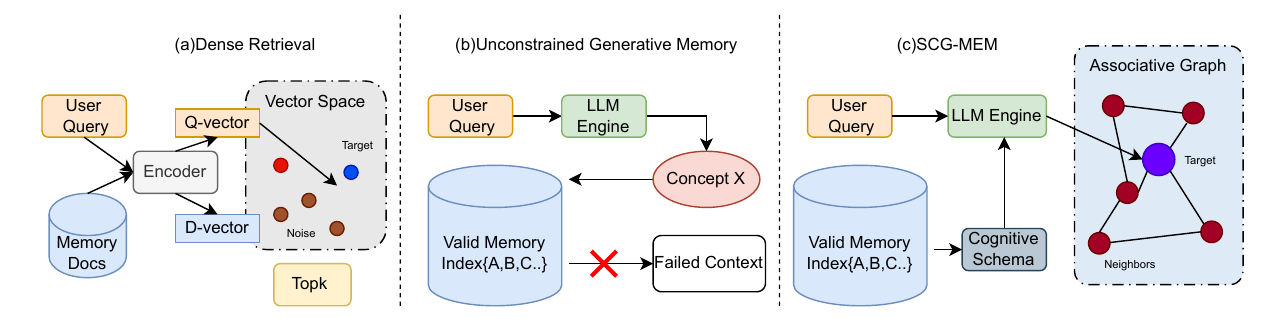}
    \caption{\textbf{Comparison of Memory Access Paradigms.} 
    \textbf{(a) Dense Retrieval:} Encodes queries and memory into vectors and retrieves top-$k$ entries via similarity matching. While structurally safe ($\hat{k} \in \mathcal{S}$), it suffers from the semantic gap where nearest neighbors may be contextually irrelevant.
    \textbf{(b) Unconstrained Generative Memory:} Directly prompts the LLM to generate memory keys. This approach risks \textit{Structural Hallucination}---producing semantically plausible but non-existent keys (e.g., ``Concept X'' $\notin \mathcal{S}$), leading to lookup failures.
    \textbf{(c) \textsc{SCG-Mem}:} Constrains LLM decoding via a \textit{Cognitive Schema} (Prefix Trie), guaranteeing that all generated keys are valid ($\hat{k} \in \mathcal{S}$). The \textit{Associative Graph} then enables multi-hop traversal to gather contextually relevant neighbors.}
    \label{fig:paradigm_shift}
\end{figure*}

\begin{figure*}[t]
    \centering
    \includegraphics[width=0.95\textwidth]{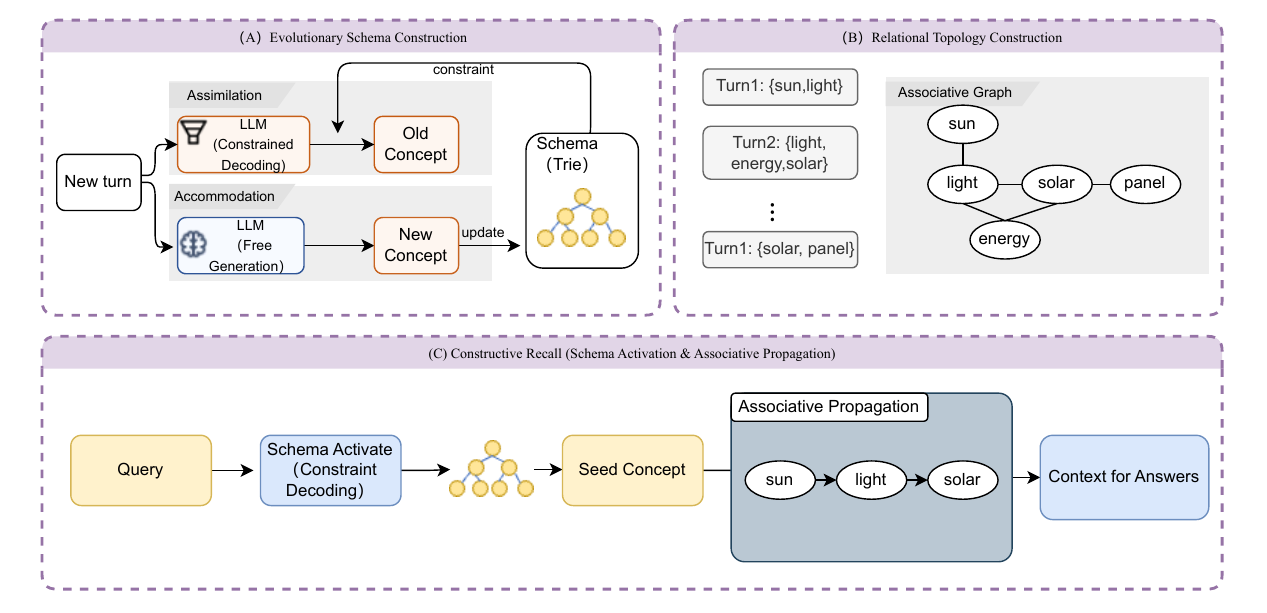}
    \caption{\textbf{The SCG-Mem Framework.} 
    \textbf{(A) Evolutionary Schema Construction:} New dialogue turns are processed via dual pathways---\textit{Assimilation} grounds inputs to existing schema nodes through constrained decoding, while \textit{Accommodation} expands the Prefix Trie with novel concepts via free generation.
    \textbf{(B) Relational Topology Construction:} Co-occurring concepts within each turn are linked in an \textit{Associative Graph}, with edge weights computed by accumulated IDF products to capture semantic coupling strength.
    \textbf{(C) Constructive Recall:} Given a query, the system first activates seed concepts via schema-constrained decoding (guaranteeing $\hat{k} \in \mathcal{S}$), then performs associative propagation over the graph to gather contextually relevant memory entries for response generation.}
    \label{fig:framework}
\end{figure*}

\section{Introduction}

Long-term memory is a fundamental capability for autonomous agents, enabling coherent reasoning, personalization, and temporal consistency across extended interactions. In recent years, architectures such as MemGPT \cite{packer2023memgpt} and RAG-based systems have proliferated \cite{zhong2024memorybank,sarthi2024raptor,lee2024human}. Despite their differences, these systems share a common empiricist assumption: memory access is a \textit{discriminative retrieval} problem. Given a query, the agent searches an external vector store to retrieve a subset of candidate entries based on approximate similarity.

While effective in short-horizon settings, this retrieval-centric paradigm exhibits persistent limitations. 
First, dense retrieval relies on identifying entities within sentences, yet identical entities frequently recur across different contexts. 
Since semantic similarity is not equivalent to contextual relevance, embeddings often fail to distinguish semantically identical but contextually distinct instances~\cite{xu2025dense}.
Second, most retrieval indices are topologically flat, lacking the 
relational structure required for associative multi-hop reasoning. 
Although recent attempts introduce graph structures~\cite{edge2024local,rezazadeh2024isolated}, 
they still rely on dense retrieval to select initial entry nodes, inheriting the same noise problem.

A natural alternative is to reformulate recall as \textit{generative reconstruction} \cite{li2025retrollm} to exploit the world knowledge and memory understanding capabilities of Large Language Models (LLMs).
 Memory access typically necessitates the construction of indices, such as inverted indices or graphs, where accessing content requires a specific \textit{key}. 
 However, allowing Large Language Models to directly generate these keys often results in the production of keys that do not exist in the memory. 
 This is a form of hallucination~\cite{huang2025survey}, we define this phenomenon as \textbf{Structural Hallucination}. 
 Such structural hallucinations are catastrophic, as they inevitably lead to lookup failures by pointing to non-existent entries.

Inspired by Piaget's constructivist epistemology~\cite{piaget1970genetic,piaget1952origins}, we recognize that human memory is structured by cognitive schemas---dynamic mental templates for understanding the world---and the act of recollection is a reconstructive process constrained by these existing structures.
Building on this insight, we propose \textbf{\textsc{SCG-Mem}} (Schema-Constrained Generative Memory), a paradigm shift that reformulates memory access from external retrieval to \textit{Schema-Constrained Generation} (Figure~\ref{fig:paradigm_shift}).

To operationalize this constructivist framework, we shift memory representation from continuous vectors to a discrete, schema-grounded construction process, where valid recall is confined by an internalized \textbf{Cognitive Schema}.
The architecture of \textsc{SCG-Mem} is built upon three synergistic components that collectively govern the existence, evolution, and association of memory.
We first distill raw memory entries into discrete \textbf{Concepts} (keywords). 
Collectively, these concepts form the agent's \textbf{Cognitive Schema}, representing the epistemic boundary of its valid knowledge. 
Technically, we structure this schema as a dynamic \textbf{Prefix Trie}. 
During recollection, this Trie functions as a hard constraint on the LLM's decoding process, ensuring that the agent only generates retrieval keys that correspond to valid paths within the schema. 
By strictly confining the search space to the schema, we mathematically preclude structural hallucinations, guaranteeing that every generated key maps to a valid memory entry.

Crucially, this schema is not static but evolves over time. 
We model the temporal dynamics through \textbf{Evolutionary Schema Construction}. 
Following Piagetian dynamics, the system continuously updates the Trie via \textit{assimilation} (grounding new inputs into existing paths) and \textit{accommodation} (expanding the Trie with novel concepts). 
Finally, to capture the semantic relationships between these dynamically evolved concepts, we overlay an \textbf{Associative Graph}. 
This transforms the schema from a discrete lexicon into a navigable cognitive map, enabling the agent to traverse associative pathways beyond explicit query matches.

We evaluate \textsc{SCG-Mem} on the LoCoMo benchmark. Experiments show that our generative-constructivist approach yields consistent and substantial improvements over retrieval-based baselines across evaluation categories.

In summary, this work makes the following contributions:
\begin{itemize}
    \item We propose \textbf{\textsc{SCG-Mem}}, a novel memory architecture that reformulates retrieval as \textit{schema-constrained generation}. By introducing constrained decoding within a valid schema, we effectively eliminate structural hallucinations.
    \item We introduce an \textbf{Associative Graph} over a Prefix-Trie-based Schema. This hybrid structure enforces key validity via the Trie while enabling associative reasoning via graph traversal.
    \item We design an \textbf{Evolutionary Schema Construction} mechanism (Assimilation and Accommodation) that enables stable yet adaptive long-term memory growth.
\end{itemize}

\section{Related Work}
\label{sec:related_work}


\subsection{Agent Memory}
Research on agent memory has explored various architectures to optimize memory organization and access. Early systems like MemoryBank~\cite{zhong2024memorybank} and MemGPT~\cite{packer2023memgpt} fragment texts into chunks managed via dense retrieval or cache-like tiers, while ReadAgent~\cite{lee2024human} utilizes gist compression for interactive lookups. To support higher-level reasoning, frameworks such as RAPTOR~\cite{sarthi2024raptor} and GraphRAG~\cite{edge2024local} structure data into recursive trees or knowledge graphs; however, these are often confined to static corpora, necessitating costly reconstruction for updates. Recent dynamic approaches like MemTree~\cite{rezazadeh2024isolated} and CAM~\cite{li2025cam} adopt Piagetian-inspired tree structures for online clustering. Yet, despite these structural advances, they fundamentally rely on discriminative retrieval for initial access, which remains vulnerable to noise from contextually irrelevant but semantically similar vectors. 
Distinctively, \textsc{SCG-Mem} departs from this retrieval paradigm by introducing schema-constrained decoding, enabling the agent to constructively accurate recall multiple entries in a single generative pass guided by its internalized cognitive schema.

\subsection{Constraint Decoding}
Constrained decoding modifies the probability distribution
 of a language model during inference to satisfy external 
 constraints.
Initial applications focused on lexical constraints, 
such as Grid Beam Search \cite{hokamp2017lexically,post2018fast},
 to ensure specific keywords appeared in the output. 
 More recently, this has evolved into syntactic 
 constraints for code generation, where systems 
 like Synchromesh\cite{poesia2022synchromesh} or PICARD\cite{scholak2021picard} enforce 
 validity against formal grammars 
 by masking invalid tokens at the logit level.
In the domain of generative retrieval, RetroLLM~\cite{li2025retrollm} employs FM-index constraints to directly generate fine-grained evidence.
Recent advances have extended constrained decoding to graph traversal tasks. 
Methods like GCR~\cite{luo2024graph} and DoG~\cite{li2025decoding} employ constraints to guide 
LLMs in selecting valid nodes during reasoning over knowledge graphs.
However, these approaches operate on static graphs and do not account for dynamic schema evolution.
\textsc{SCG-Mem} adapts this mechanism from the 
syntactic to the \textit{semantic} 
and \textit{ontological} domain. 
Instead of enforcing valid syntax, we enforce 
valid memory paths via a Prefix Trie. Unlike 
standard constrained decoding which assumes a 
static constraint set (e.g., a fixed grammar), 
our approach introduces \textit{dynamic} constraints 
that evolve through the agent's lifetime.

\section{Methodology}
\label{sec:methodology}

We present \textbf{\textsc{SCG-Mem}} (Schema-Constrained Generative Memory), a memory architecture grounded in the constructivist principle that knowing is an active process of construction under constraints. Unlike traditional retrieval systems that treat memory as a static repository accessed via similarity search, \textsc{SCG-Mem} reformulates memory access as a generative process governed by a dynamic cognitive schema.

As illustrated in Figure \ref{fig:framework}, the framework is composed of three tightly coupled components working in synergy. At the foundation lies a \textbf{Cognitive Schema} (implemented as a Prefix Trie), which defines the epistemic boundary of the agent and strictly enforces valid memory access (Section \ref{subsec:problem_formulation} \& \ref{subsec:schema}). Crucially, this schema is not static; we employ an \textbf{Evolutionary Schema Construction} mechanism that dynamically updates the schema through \textit{assimilation} and \textit{accommodation} processes to ensure long-term adaptability (Section \ref{subsec:evolution}). Finally, to support reasoning capabilities beyond simple validity, we overlay this evolving schema with an \textbf{Associative Graph} that enables associative multi-hop reasoning via activation propagation (Section \ref{subsec:topology}). In the following sections, we first formally define the problem of Structural Hallucination, and then detail how each component addresses it to achieve robust, evolving memory.

\subsection{Problem Formulation: Structural Hallucination}
\label{subsec:problem_formulation}

In the context of autonomous agents, memory access is typically modeled as a mapping from a query context $c$ to a memory entry key $k$. We define the agent's cognitive schema $\mathcal{S}$ as a finite set of valid concept keys.

\begin{definition}[\textbf{Structural Hallucination}]
Given a context $c$ and a schema $\mathcal{S}$, a generated memory entry point $\hat{k}$ is a Structural Hallucination if and only if:
\begin{equation}
    \hat{k} \notin \mathcal{S}
\end{equation}
even if $\hat{k}$ is semantically plausible or factually correct in a broad world knowledge context.
\end{definition}

\noindent\textbf{Remark.}· Structural hallucination differs fundamentally from retrieval noise: retrieval systems may return irrelevant keys, but always valid ones ($\hat{k} \in \mathcal{S}$), whereas generative models can produce non-existent keys that cause lookup failures.

Our objective is to construct a memory access mechanism $P_{\theta}$ such that the probability of structural hallucination is strictly zero:
\begin{equation}
    P_{\theta}(\hat{k} \notin \mathcal{S} | c) = 0
\end{equation}

\subsection{The Cognitive Schema}
\label{subsec:schema}

To enforce the validity constraint established in Eq. (2), we explicitly model the agent's epistemic boundaries. We operationalize the cognitive schema $\mathcal{S}$ not as a static database, but as a dynamic structural constraint implemented via a \textbf{Prefix Trie} $\mathcal{T}$.

\begin{definition}[\textbf{Cognitive Schema}]
\label{def:schema}
The Cognitive Schema $\mathcal{S}$ is formally defined as a finite set of concept keys over the token vocabulary $\Sigma$:
\begin{equation}
    \mathcal{S} \subset \Sigma^*
\end{equation}
where $\Sigma^*$ denotes the set of all finite token sequences.
To make this set computationally enforceable during generation, we construct a Prefix Trie $\mathcal{T}$ such that every path from the root to a marked end-node corresponds to a valid key $k \in \mathcal{S}$.

Mathematically, $\mathcal{T}$ defines the prefix-closed validity space $\Omega_{\mathcal{S}}$ within the universe of all token sequences $\Sigma^*$:
\begin{equation}
    \Omega_{\mathcal{S}} = \{ s \in \Sigma^* \mid \exists k \in \mathcal{S}, s \text{ is a prefix of } k \}
\end{equation}
A generated memory entry $\hat{k}$ satisfies the non-hallucination condition (Eq. 2) if and only if $\hat{k} \in \mathcal{S} \subset \Omega_{\mathcal{S}}$.
\end{definition}

Crucially, the schema functions as a \textit{hard constraint} on the language model's decoding manifold. We define a binary validity indicator $\mathbb{I}_{\mathcal{S}}(y_{1:t})$ for any generated token sequence $y_{1:t}$ at step $t$:
\begin{equation}
    \mathbb{I}_{\mathcal{S}}(y_{1:t}) = 
    \begin{cases} 
    1 & \text{if } y_{1:t} \in \Omega_{\mathcal{S}} \\
    0 & \text{otherwise}
    \end{cases}
\end{equation}
This indicator acts as a gatekeeper, pruning the probability mass of invalid tokens to zero before the softmax layer (detailed in Sec. \ref{subsec:decoding}).

By adopting this architecture, we achieve a fundamental dissociation between \textit{existence} and \textit{association}:
\begin{itemize}
    \item \textbf{Existence (Ontology):} Managed by the Trie $\mathcal{T}$. It answers the question \textit{``Does this concept exist in my world?''} by strictly enforcing $\hat{k} \in \mathcal{S}$.
    \item \textbf{Association (Topology):} Managed by the Associative Graph. It answers \textit{``How is this concept related to others?''} via edge weights (Sec. \ref{subsec:topology}).
\end{itemize}
This separation guarantees that the agent's reasoning process is strictly confined to its valid epistemic boundaries, thereby eliminating structural hallucinations by construction.

\subsection{Schema-Constrained Decoding}
\label{subsec:decoding}

Memory access in \textsc{SCG-Mem} is realized through schema-constrained decoding. Given a context $c$, we modify the token-level generation process of the LLM (parameterized by $\theta$) such that the probability mass is strictly confined to the validity space $\Omega_{\mathcal{S}}$.

Formally, let $y_{<t}$ be the generated prefix at step $t$. The schema-constrained distribution $P_{\mathcal{S}}$ is defined as the renormalization of the base model distribution $P_{\theta}$ masked by the validity indicator $\mathbb{I}_{\mathcal{S}}$ (defined in Eq. 4):

\begin{equation}
    P_{\mathcal{S}}(y_t | y_{<t}, c) = \frac{P_{\theta}(y_t | y_{<t}, c) \cdot \mathbb{I}_{\mathcal{S}}(y_{<t} \circ y_t)}{Z(y_{<t})}
\end{equation}

where $\circ$ denotes string concatenation and $Z(y_{<t})$ is the normalization constant. 
This mechanism enforces an \emph{ex-ante epistemic constraint}: during autoregressive decoding, 
any token $y_t$ that results in a prefix outside $\Omega_{\mathcal{S}}$ receives zero probability.
Consequently, the agent is mathematically incapable of generating a completed key $\hat{k}$ such
 that $\hat{k} \notin \mathcal{S}$.

\subsection{Schema Evolution: Assimilation and Accommodation}
\label{subsec:evolution}

Because the schema $\mathcal{S}$ strictly constrains generation, it must distinguish between ``known concepts'' and ``novel information.'' We model this via \textbf{Evolutionary Schema Construction}, a dual-pathway process inspired by Piagetian dynamics. Let $\mathcal{S}^{(t-1)}$ denote the concept set at time $t-1$. When a new interaction $d^{(t)}$ arrives:

\paragraph{Assimilation (Grounding).}
The agent first attempts to interpret the input using its existing cognitive structures. We perform schema-constrained generation to map the input to valid keys within the current ontology:
\begin{equation}
    \mathcal{K}_{\text{assim}} = \{ k \in \mathcal{S}^{(t-1)} \mid k \sim P_{\mathcal{S}}(\cdot \mid d^{(t)}) \}
\end{equation}
Assimilation reinforces the weights of existing graph edges without altering the epistemic boundary $\mathcal{S}$.

\paragraph{Accommodation (Expansion).}
If constrained generation fails to adequately represent the input (e.g., measured by high perplexity or a specific "unknown" token), the system triggers accommodation. The schema constraint is temporarily relaxed, allowing the base model $P_{\theta}$ to generate novel concepts:
\begin{equation}
    \mathcal{K}_{\text{nov}} = \{ k \mid k \sim P_{\theta}(\cdot \mid d^{(t)}) \} \setminus \mathcal{S}^{(t-1)}
\end{equation}
These novel concepts are then validated and inserted into the Trie, explicitly expanding the agent's epistemic boundary:
\begin{equation}
    \mathcal{S}^{(t)} \leftarrow \mathcal{S}^{(t-1)} \cup \mathcal{K}_{\text{nov}}
\end{equation}
This evolutionary mechanism ensures that the memory system remains open-ended while maintaining structural consistency.

\subsection{Associative Reasoning within the Schema}
\label{subsec:topology}

While the Cognitive Schema $\mathcal{S}$ enforces \textit{existence} (validity), it implies a flat topology. To support multi-hop reasoning, we overlay an \textbf{Associative Graph} modeled as a weighted undirected graph $\mathcal{G} = (V, E)$, where the vertex set $V = \mathcal{S}$ corresponds exactly to the valid concepts defined in the schema.

We define the edge weight $w_{uv}$ between two concepts $u, v \in \mathcal{S}$ based on their semantic coupling strength across the interaction history $\mathcal{H}$. 
By abstracting away linear token distance, we focus purely on the information-theoretic value of their co-occurrence. 
The weight is computed as the accumulated IDF product:

\begin{equation}
    w_{uv} = \sum_{(u,v) \in \mathcal{H}} \text{IDF}(u) \cdot \text{IDF}(v)
\end{equation}

where the Inverse Document Frequency is defined as $\text{IDF}(k) = \log \frac{N}{df(k)}$, with $N$ denoting the total number of dialogue turns and $df(k)$ the number of turns containing concept $k$. Here $\mathcal{H}$ represents the set of co-occurring concept pairs within the same dialogue turn. The $\text{IDF}(\cdot)$ term acts as a significance filter, penalizing ubiquitous stop-words while boosting the connection strength between rare, domain-specific concepts. This topology enables the propagation of activation from explicit query terms to implicitly related concepts based on accumulated semantic relevance.

\subsection{Constructive Recall}
\label{subsec:recall}

Recall in \textsc{SCG-Mem} is a constructive, dual-stage process comprising schema grounding and activation propagation.

\paragraph{Schema Activation.}
Given a query $q$, the agent first identifies valid entry points within the schema $\mathcal{S}$ for subsequent graph traversal. 
Instead of relying on fuzzy vector retrieval, we perform schema-constrained generation to obtain a set of seed concepts $K_{\text{seed}}$. 
To decode multiple diverse yet relevant keywords, we employ \textbf{constrained beam search} with beam size $b$. At each decoding step, we maintain the top-$b$ partial sequences ranked by cumulative log-probability, while enforcing the schema constraint (Eq. 5) to prune invalid branches. Upon completion, this yields $b$ distinct valid keys:
\begin{equation}
    K_{\text{seed}} = \text{BeamSearch}_b(P_{\mathcal{S}}(\cdot \mid q)), \quad |K_{\text{seed}}| = b
\end{equation}
This approach leverages the LLM's semantic understanding to identify the most query-relevant concepts while the Trie constraint guarantees that all returned keys are valid schema nodes.

\paragraph{Associative Propagation.}
To recover implicit context, we propagate activation from the seed concepts to their neighbors via weighted sampling. 
For each active node $u \in K_{\text{seed}}$, we compute the transition probability to a neighbor $v$ by applying a softmax over the edge weights $w_{uv}$ (defined in Sec. \ref{subsec:topology}):
\begin{equation}
    P(v \mid u) = \text{Softmax}(w_{uv}) = \frac{\exp(w_{uv} / T)}{\sum_{v' \in \mathcal{N}(u)} \exp(w_{uv'} / T)}
\end{equation}
where $\mathcal{N}(u)$ denotes the neighbors of $u$, and $T$ is a temperature parameter controlling the breadth of association. The set of retrieved context concepts $K_{\text{context}}$ is then obtained by sampling from this distribution:
\begin{equation}
    K_{\text{context}} \sim P(\cdot \mid K_{\text{seed}})
\end{equation}
This stochastic approach allows the agent to efficiently explore the neighborhood of valid concepts, capturing relevant details that are topologically close (high semantic coupling) to the query.


\paragraph{Context Reconstruction.}
To ground the agent's reasoning in concrete evidence, we map the activated conceptual subgraph back to the original source text. We retrieve the text segments associated with the union of seed and expanded concepts ($K_{\text{seed}} \cup K_{\text{context}}$) to form the context $C$. The final response is synthesized by the LLM, conditioned on this schema-grounded context: $r \sim P_{\theta}(r \mid C, q)$.

\section{Experiments}
\label{sec:experiments}

To rigorously evaluate \textsc{SCG-Mem}, we follow the established protocols of the \textbf{LoCoMo} benchmark \cite{maharana2024evaluating}, which represents the current state-of-the-art for evaluating long-term, multi-session agentic memory.

\subsection{Dataset}

We utilize the LoCoMo dataset~\cite{maharana2024evaluating}, featuring ultra-long dialogues (avg. 9K tokens, up to 35 sessions). We evaluate on four categories: \textbf{Single-Hop} (single-session retrieval), \textbf{Multi-Hop} (cross-session synthesis), \textbf{Temporal} (time-dependent updates), and \textbf{Adversarial} (misleading queries), excluding Open-Domain to focus on conversation-grounded memory.

\subsection{Metrics}

For evaluation, we employ two primary metrics: the \textbf{F1 Score} to assess answer accuracy by balancing precision and recall, and \textbf{BLEU-1} to evaluate generated response quality by measuring word overlap with ground truth responses.

\subsection{Baselines}

We compare \textsc{SCG-Mem} against five representative memory systems:
\textbf{LoCoMo}~\cite{maharana2024evaluating} directly leverages foundation models without memory mechanisms, incorporating the complete preceding conversation into the prompt for each query.
\textbf{ReadAgent}~\cite{lee2024human} processes long-context documents through episode pagination, memory gisting, and interactive look-up.
\textbf{MemoryBank}~\cite{zhong2024memorybank} maintains historical interactions with a dynamic memory updating mechanism based on the Ebbinghaus Forgetting Curve theory.
\textbf{MemGPT}~\cite{packer2023memgpt} implements a dual-tier virtual context management system inspired by OS memory hierarchies, with main context (RAM) and external context (disk).
\textbf{A-MEM}~\cite{xu2025mem} reformulates raw interactions into structured ``notes'' and employs an evolutionary mechanism to establish inter-memory connections, using dense retrieval for initial node access and associative traversal for context gathering.

\subsection{Implementation Details}

Since \textsc{SCG-Mem} requires direct access to token-level probability distributions for schema-constrained decoding, 
we deploy all models locally using the Hugging Face Transformers library. 
We primarily evaluate on \textbf{Qwen2.5 (1.5B, 3B)} \cite{hui2024qwen2} and \textbf{Llama 3.2 (1B, 3B)} \cite{grattafiori2024llama}, instantiated with full precision for accurate logit manipulation. 
For text embedding, we utilize the \textbf{BGE-M3} \cite{chen2024bge} model across all experiments to construct the initial concept representations. 

\begin{table*}[t]
\centering
\caption{Experimental results on the LoCoMo benchmark across different foundation models. Since \textsc{SCG-Mem} requires access to token-level probability distributions for schema-constrained decoding, we evaluate on open-weights models (Qwen 2.5 and Llama 3.2). \textsc{SCG-Mem} consistently outperforms the strongest baseline A-MEM, demonstrating the efficacy of structural constraints even on smaller-scale models. (Best results in \textbf{bold}).}
\resizebox{\textwidth}{!}{
\begin{tabular}{c|l|cc|cc|cc|cc|cc}
\toprule
\multirow{2}{*}{\textbf{Model}} & \multirow{2}{*}{\textbf{Method}} & \multicolumn{2}{c|}{\textbf{Multi Hop}} & \multicolumn{2}{c|}{\textbf{Temporal}} & \multicolumn{2}{c|}{\textbf{Single Hop}} & \multicolumn{2}{c|}{\textbf{Adversarial}} & \multicolumn{2}{c}{\textbf{Average}} \\
 & & \textbf{F1} & \textbf{BLEU} & \textbf{F1} & \textbf{BLEU} & \textbf{F1} & \textbf{BLEU} & \textbf{F1} & \textbf{BLEU} & \textbf{F1} & \textbf{BLEU} \\
\midrule

\multirow{6}{*}{\rotatebox[origin=c]{90}{\textbf{Qwen2.5 1.5B}}} 
 & LoCoMo & 9.05 & 6.55 & 4.25 & 4.04 & 11.15 & 8.67 & 40.38 & 40.23 & 16.21 & 14.87 \\
 & ReadAgent & 6.61 & 4.93 & 2.55 & 2.51 & 10.13 & 7.54 & 5.42 & 27.32 & 6.18 & 10.58 \\
 & MemoryBank & 11.14 & 8.25 & 4.46 & 2.87 & 13.42 & 11.01 & 36.76 & 34.00 & 16.45 & 14.03 \\
 & MemGPT & 10.44 & 7.61 & 4.21 & 3.89 & 9.56 & 7.34 & 31.51 & 28.90 & 13.93 & 11.94 \\
 & A-MEM & 18.23 & 11.94 & 24.32 & 19.74 & 23.63 & 19.23 & 46.00 & 43.26 & 28.05 & 23.54 \\
 & \cellcolor{gray!15}\textbf{SCG-Mem (Ours)} & \textbf{20.06} & \textbf{16.22} & \textbf{28.33} & \textbf{21.16} & \textbf{31.18} & \textbf{21.34} & \textbf{71.99} & \textbf{68.21} & \textbf{37.89} & \textbf{31.73} \\
\cmidrule{1-12}
\multicolumn{2}{c|}{\textit{Improvement (\%)}} & \textit{+10.0\%} & \textit{+35.8\%} & \textit{+16.5\%} & \textit{+7.2\%} & \textit{+32.0\%} & \textit{+11.0\%} & \textit{+56.5\%} & \textit{+57.7\%} & \textit{+35.1\%} & \textit{+34.8\%} \\
\midrule

\multirow{6}{*}{\rotatebox[origin=c]{90}{\textbf{Qwen2.5 3B}}} 
 & LoCoMo & 4.61 & 4.29 & 3.11 & 2.71 & 7.03 & 5.69 & 16.95 & 14.81 & 7.93 & 6.88 \\
 & ReadAgent & 2.47 & 1.78 & 3.01 & 3.01 & 3.25 & 2.51 & 15.78 & 14.01 & 6.13 & 5.33 \\
 & MemoryBank & 3.60 & 3.39 & 1.72 & 1.97 & 4.11 & 3.32 & 13.07 & 10.30 & 5.63 & 4.75 \\
 & MemGPT & 5.07 & 4.31 & 2.94 & 2.95 & 7.26 & 5.52 & 14.47 & 12.39 & 7.44 & 6.29 \\
 & A-MEM & 12.57 & 9.01 & 27.59 & 25.07 & 17.23 & 13.12 & 27.91 & 25.15 & 21.33 & 18.09 \\
 & \cellcolor{gray!15}\textbf{SCG-Mem (Ours)} & \textbf{28.49} & \textbf{21.21} & \textbf{42.29} & \textbf{31.72} & \textbf{42.51} & \textbf{32.14} & \textbf{52.63} & \textbf{48.76} & \textbf{41.48} & \textbf{33.46} \\
\cmidrule{1-12}
\multicolumn{2}{c|}{\textit{Improvement (\%)}} & \textit{+126.6\%} & \textit{+135.4\%} & \textit{+53.3\%} & \textit{+26.5\%} & \textit{+146.7\%} & \textit{+145.0\%} & \textit{+88.6\%} & \textit{+93.9\%} & \textit{+94.5\%} & \textit{+85.0\%} \\
\midrule

\multirow{6}{*}{\rotatebox[origin=c]{90}{\textbf{Llama 3.2 1B}}} 
 & LoCoMo & 11.25 & 9.18 & 7.38 & 6.82 & 12.86 & 10.50 & 51.89 & 48.27 & 20.85 & 18.69 \\
 & ReadAgent & 5.96 & 5.12 & 1.93 & 2.30 & 7.75 & 6.03 & 44.64 & 40.15 & 15.07 & 13.40 \\
 & MemoryBank & 13.18 & 10.03 & 7.61 & 6.27 & 17.30 & 14.03 & 52.61 & 47.53 & 22.68 & 19.47 \\
 & MemGPT & 9.19 & 6.96 & 4.02 & 4.79 & 10.16 & 7.68 & 49.75 & 45.11 & 18.28 & 16.14 \\
 & A-MEM & 19.06 & 11.71 & 17.80 & 10.28 & 28.51 & 24.13 & 58.81 & 54.28 & 31.05 & 25.10 \\
 & \cellcolor{gray!15}\textbf{SCG-Mem (Ours)} & \textbf{36.66} & \textbf{29.06} & \textbf{37.64} & \textbf{24.92} & \textbf{41.19} & \textbf{32.17} & \textbf{63.21} & \textbf{58.69} & \textbf{44.68} & \textbf{36.21} \\
\cmidrule{1-12}
\multicolumn{2}{c|}{\textit{Improvement (\%)}} & \textit{+92.3\%} & \textit{+148.2\%} & \textit{+111.5\%} & \textit{+142.4\%} & \textit{+44.5\%} & \textit{+33.3\%} & \textit{+7.5\%} & \textit{+8.1\%} & \textit{+43.9\%} & \textit{+44.3\%} \\
\midrule

\multirow{6}{*}{\rotatebox[origin=c]{90}{\textbf{Llama 3.2 3B}}} 
 & LoCoMo & 6.88 & 5.77 & 4.37 & 4.40 & 8.37 & 6.93 & 30.25 & 28.46 & 12.47 & 11.39 \\
 & ReadAgent & 2.47 & 1.78 & 3.01 & 3.01 & 3.25 & 2.51 & 15.78 & 14.01 & 6.13 & 5.33 \\
 & MemoryBank & 6.19 & 4.47 & 3.49 & 3.13 & 7.61 & 6.03 & 18.65 & 17.05 & 8.99 & 7.67 \\
 & MemGPT & 5.32 & 3.99 & 2.68 & 2.72 & 4.32 & 3.51 & 21.45 & 19.37 & 8.44 & 7.40 \\
 & A-MEM & 17.44 & 11.74 & 26.38 & 19.50 & 28.14 & 23.87 & 42.04 & 40.60 & 28.50 & 23.93 \\
 & \cellcolor{gray!15}\textbf{SCG-Mem (Ours)} & \textbf{20.16} & \textbf{12.67} & \textbf{49.35} & \textbf{34.27} & \textbf{44.39} & \textbf{37.71} & \textbf{49.30} & \textbf{48.23} & \textbf{40.80} & \textbf{33.22} \\
\cmidrule{1-12}
\multicolumn{2}{c|}{\textit{Improvement (\%)}} & \textit{+15.6\%} & \textit{+7.9\%} & \textit{+87.1\%} & \textit{+75.7\%} & \textit{+57.7\%} & \textit{+58.0\%} & \textit{+17.3\%} & \textit{+18.8\%} & \textit{+43.2\%} & \textit{+38.8\%} \\
\bottomrule
\end{tabular}
}
\label{tab:local_models_comparison}
\end{table*}

\subsection{Main Results}
\label{subsec:main_results}

Table \ref{tab:local_models_comparison} shows \textsc{SCG-Mem} consistently outperforms the strongest baseline A-MEM across all categories on both Qwen 2.5 and Llama 3.2 models, with notable gains exceeding \textbf{100\%} in specific tasks (e.g., \textbf{+146.7\%} F1 in \textit{Single-Hop} on Qwen2.5 3B).
Specifically, \textit{Multi-Hop} reasoning sees up to \textbf{+126.6\%} F1 improvement on Qwen2.5 3B, validating that our hybrid architecture---combining Trie-based constrained decoding with Associative Graph traversal---successfully bridges disjoint contexts that challenge standard retrieval. 
Concurrently, \textit{Temporal} tasks show substantial gains (e.g., \textbf{+111.5\%} F1 on Llama 3.2 1B), confirming that the assimilation-accommodation mechanism effectively updates epistemic boundaries to track evolving facts.
Finally, in \textit{Adversarial} settings—deliberately designed to mislead with traps and uncertainty—\textsc{SCG-Mem} demonstrates that such queries can be accurately mapped to relevant memories through semantic mapping, boosting performance by up to \textbf{+88.6\%} F1 on Qwen2.5 3B.

\subsection{Ablation Study}
\label{subsec:ablation}

To analyze the contribution of each component, we conduct an ablation study using Qwen 2.5 3B as the foundation model across all four task categories (Table \ref{tab:ablation}).
\paragraph{w/o Cognitive Constraint.}
We replace schema-constrained decoding with unconstrained generation, then concatenate the generated keywords into a single string and use its embedding to retrieve seed concepts from the Cognitive Schema via dense similarity matching. 
Results show severe degradation, notably \textit{Multi-Hop} F1 dropping by \textbf{-39.5\%}. 
This confirms that accurate seed concept selection is critical for effective memory recall.

\paragraph{w/o Evolutionary Update.}
We disable assimilation (grounding to existing schema nodes) and retain only accommodation (generating novel concepts). 
This inability to connect new information with existing knowledge causes substantial drops: \textit{Multi-Hop} F1 decreases by \textbf{-34.2\%} and \textit{Temporal} by \textbf{-20.1\%}. 
This highlights that linking new inputs to existing epistemic structures is essential for bridging disjoint contexts and tracking evolving facts.

\begin{table}[t]
\centering
\caption{Ablation study on key components across all four task categories. We evaluate the impact of removing the schema constraint and the evolutionary mechanism. The results confirm that each component is essential for its specific target capability (e.g., Schema for Adversarial, Evolution for Temporal).}
\resizebox{\columnwidth}{!}{
\begin{tabular}{l|cc|cc|cc|cc}
\toprule
\multirow{2}{*}{\textbf{Configuration}} & \multicolumn{2}{c|}{\textbf{Multi-Hop}} & \multicolumn{2}{c|}{\textbf{Temporal}} & \multicolumn{2}{c|}{\textbf{Single Hop}} & \multicolumn{2}{c}{\textbf{Adversarial}} \\
 & \textbf{F1} & \textbf{BLEU} & \textbf{F1} & \textbf{BLEU} & \textbf{F1} & \textbf{BLEU} & \textbf{F1} & \textbf{BLEU} \\
\midrule
\textbf{SCG-Mem (Full)} & \textbf{28.49} & \textbf{21.21} & \textbf{42.29} & \textbf{31.72} & \textbf{42.51} & \textbf{32.14} & \textbf{52.63} & \textbf{48.76} \\
\midrule

\textit{w/o Cognitive Constraint} & 17.23 & 14.38 & 35.29 & 24.92 & 34.45 & 24.87 & 41.74 & 39.79 \\
\multicolumn{1}{r|}{\scriptsize \textit{\textcolor{gray}{Performance Drop}}} & \scriptsize \textit{-39.5\%} & \scriptsize \textit{-32.2\%} & \scriptsize \textit{-16.6\%} & \scriptsize \textit{-21.4\%} & \scriptsize \textit{-19.0\%} & \scriptsize \textit{-22.6\%} & \scriptsize \textit{-20.7\%} & \scriptsize \textit{-18.4\%} \\
\midrule

\textit{w/o Evolutionary Update} & 18.74 & 14.24 & 33.79 & 24.59 & 33.73 & 25.70 & 34.33 & 31.75 \\
\multicolumn{1}{r|}{\scriptsize \textit{\textcolor{gray}{Performance Drop}}} & \scriptsize \textit{-34.2\%} & \scriptsize \textit{-32.9\%} & \scriptsize \textit{-20.1\%} & \scriptsize \textit{-22.5\%} & \scriptsize \textit{-20.7\%} & \scriptsize \textit{-20.0\%} & \scriptsize \textit{-34.8\%} & \scriptsize \textit{-34.9\%} \\
\bottomrule
\end{tabular}
}
\label{tab:ablation}
\end{table}

\subsection{Hyperparameter Sensitivity}
\label{subsec:hyperparameter}

We examine the impact of two critical hyperparameters: the number of retrieved concepts $k$ for context reconstruction, and the depth of associative propagation (hops) on the graph topology.

\begin{figure}[t]
    \centering
    \begin{minipage}{0.48\columnwidth}
        \centering
        \includegraphics[width=\linewidth]{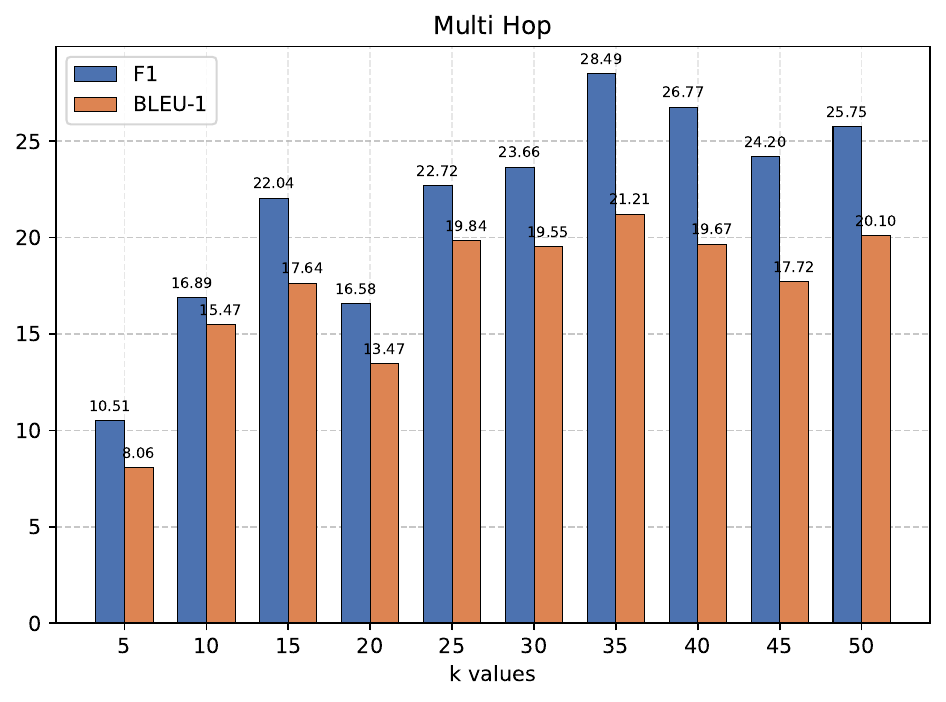}
        \centerline{\small (a) Multi-Hop}
    \end{minipage}
    \hfill
    \begin{minipage}{0.48\columnwidth}
        \centering
        \includegraphics[width=\linewidth]{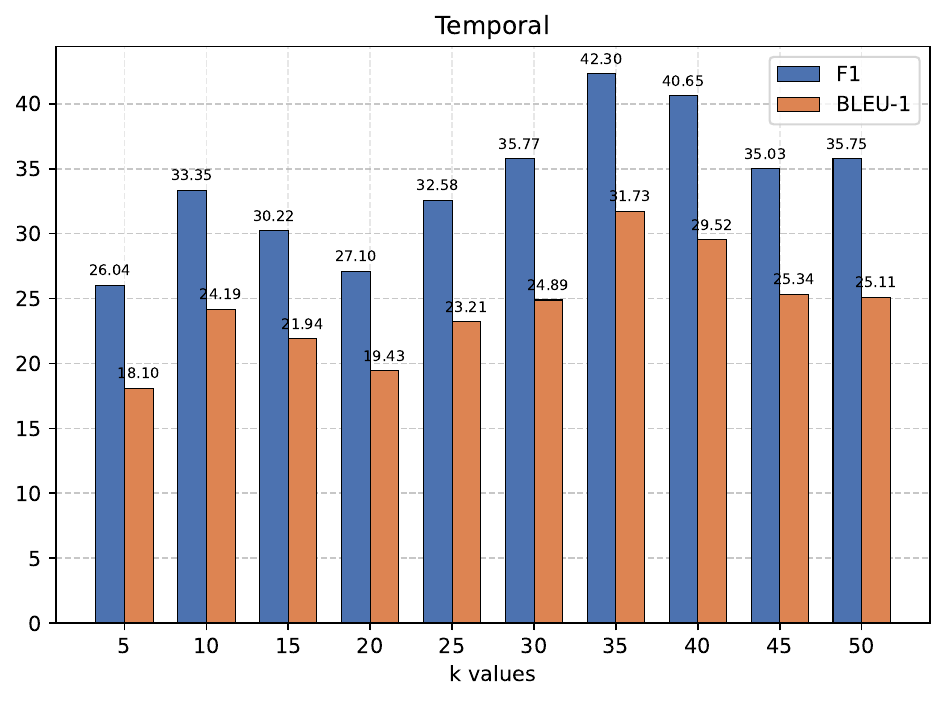}
        \centerline{\small (b) Temporal}
    \end{minipage}
    
    \vspace{2mm}

    \begin{minipage}{0.48\columnwidth}
        \centering
        \includegraphics[width=\linewidth]{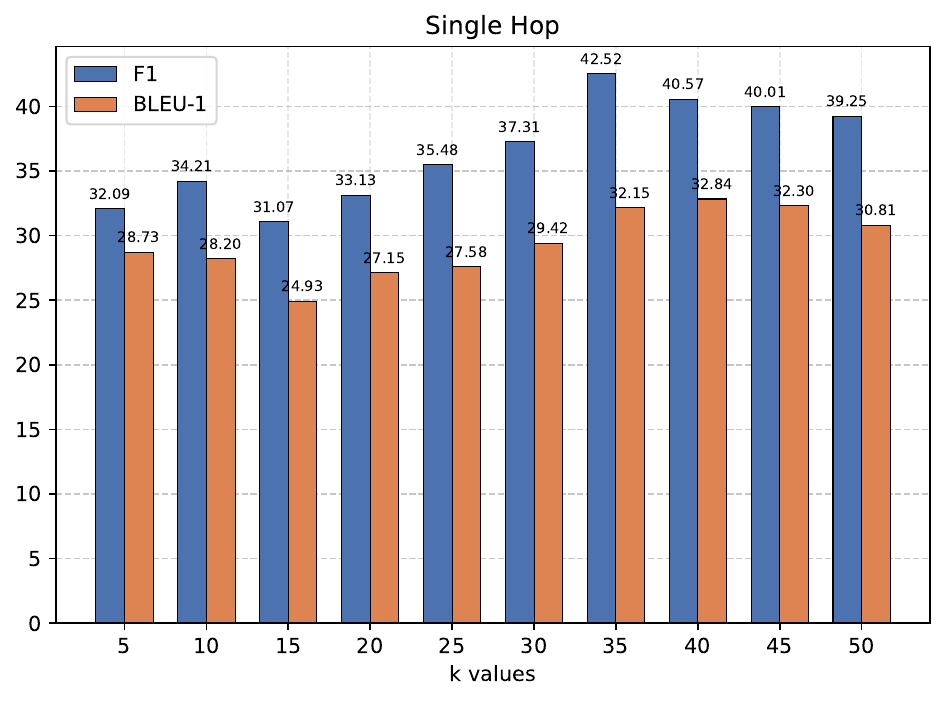}
        \centerline{\small (c) Single-Hop}
    \end{minipage}
    \hfill
    \begin{minipage}{0.48\columnwidth}
        \centering
        \includegraphics[width=\linewidth]{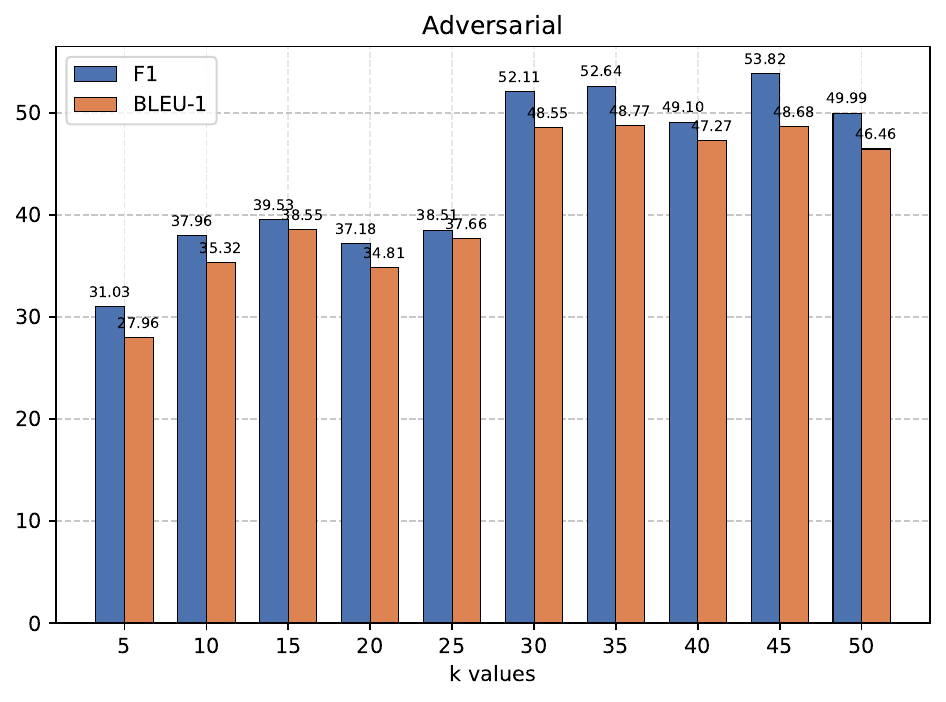}
        \centerline{\small (d) Adversarial}
    \end{minipage}
    
    \caption{\textbf{Impact of Retrieval Size $k$.} Performance exhibits a consistent rise-then-fall pattern across all categories, peaking around $k=35$. Insufficient retrieval ($k<20$) misses relevant context, while excessive retrieval ($k>40$) introduces noise that degrades reasoning precision.}
    \label{fig:hyperparameters}
\end{figure}

\paragraph{Impact of Retrieval Size ($k$).}
As illustrated in Figure \ref{fig:hyperparameters}, performance follows a characteristic inverted-U curve across all task categories, peaking around $k=35$. In the ascending phase ($k<35$), increasing retrieval size progressively enriches the context with relevant concepts, enabling more comprehensive reasoning. Beyond the optimal point, performance gradually declines as excessive retrieval introduces semantically related but contextually irrelevant information, which dilutes the signal and impairs the LLM's reasoning precision. Notably, complex tasks (Multi-Hop and Temporal) exhibit steeper gains during the ascending phase, reflecting their greater dependence on sufficient contextual coverage.

\begin{figure}[t]
    \centering
    \begin{minipage}{0.48\columnwidth}
        \centering
        \includegraphics[width=\linewidth]{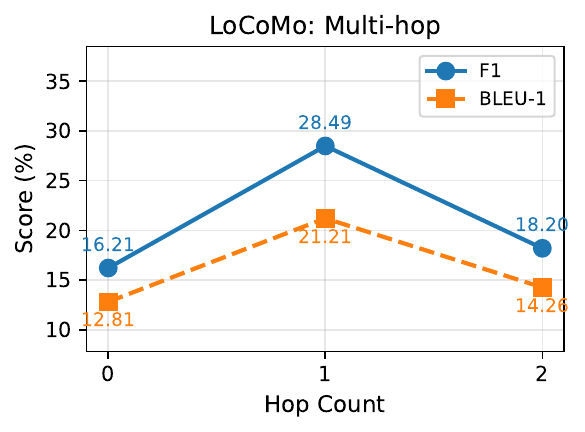}
        \centerline{\small (a) Multi-Hop}
    \end{minipage}
    \hfill
    \begin{minipage}{0.48\columnwidth}
        \centering
        \includegraphics[width=\linewidth]{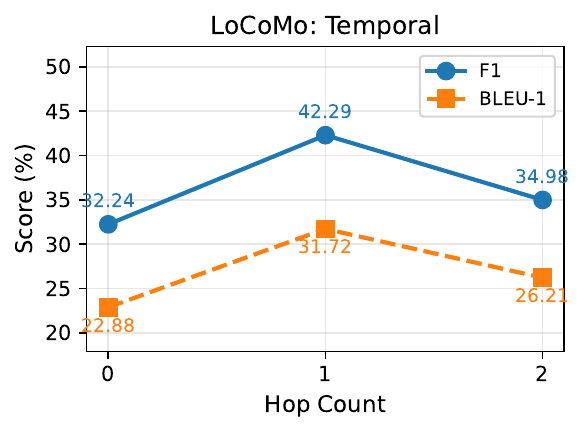}
        \centerline{\small (b) Temporal}
    \end{minipage}
    
    \vspace{2mm}

    \begin{minipage}{0.48\columnwidth}
        \centering
        \includegraphics[width=\linewidth]{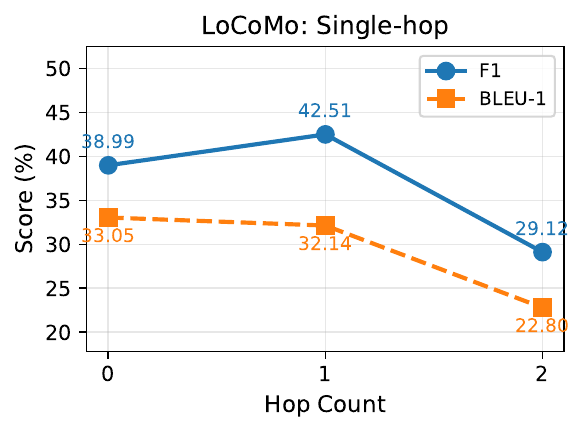}
        \centerline{\small (c) Single-Hop}
    \end{minipage}
    \hfill
    \begin{minipage}{0.48\columnwidth}
        \centering
        \includegraphics[width=\linewidth]{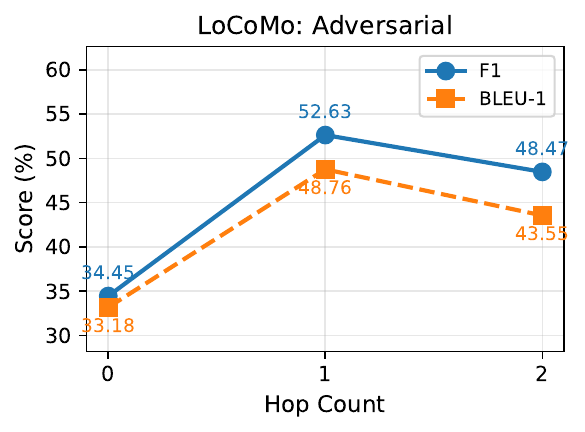}
        \centerline{\small (d) Adversarial}
    \end{minipage}
    
    \caption{\textbf{Impact of Hop Count across Categories.} Performance consistently peaks at hop-1 across all categories, demonstrating the critical value of one-step associative propagation. Multi-Hop exhibits the largest relative gain from hop-0 to hop-1, as direct schema matching alone cannot bridge disjoint conversation sessions. However, hop-2 uniformly degrades performance, with Temporal reasoning suffering the most severe decline, indicating that excessive propagation introduces semantically drifted noise.}
    \label{fig:hop_analysis}
\end{figure}

\paragraph{Impact of Association Hops.}
Figure \ref{fig:hop_analysis} reveals a consistent inverted-V pattern across all task categories. Hop-0 retrieval, which relies solely on directly matched seed concepts via the Schema, yields the lowest performance---particularly on Multi-Hop tasks where information is scattered across disjoint sessions. Extending to hop-1 substantially improves all categories by recovering implicitly connected concepts through one-step graph traversal. However, hop-2 propagation universally degrades performance, with Temporal reasoning showing the steepest decline. This suggests that while shallow associative propagation effectively bridges semantic gaps, deeper traversal introduces semantically drifted concepts that dilute the context and impair reasoning precision. The optimal hop depth of 1 reflects a balance between coverage and noise.

\section{Conclusion and Future Work}
\label{sec:conclusion}

In this paper, we proposed \textsc{SCG-Mem}, a novel memory architecture that reformulates memory access from discriminative retrieval to \textbf{schema-constrained generation}. By maintaining a dynamic Prefix Trie as the cognitive schema and constraining LLM decoding to generate only valid memory entry keys, we provide a formal guarantee that eliminates structural hallucinations by construction. To support long-term adaptation, we model memory updates via Piagetian assimilation (grounding into existing schema) and accommodation (schema expansion with novel concepts). Furthermore, we construct an Associative Graph over the schema and perform activation propagation for multi-hop reasoning. Experimental results on the LoCoMo benchmark demonstrate that \textsc{SCG-Mem} consistently outperforms state-of-the-art baselines, particularly in multi-hop and adversarial tasks. 
Future work will explore \textbf{memory compression} mechanisms during storage to reduce redundancy and improve efficiency, as well as \textbf{memory rewriting} strategies that enable the agent to consolidate and refine existing knowledge over time. We also plan to investigate \textbf{hierarchical schema structures} that organize concepts at multiple levels of abstraction, 
potentially improving both retrieval efficiency and semantic coherence. 
Additionally, 
extending \textsc{SCG-Mem} to \textbf{multi-modal settings}---where the cognitive schema encompasses visual, 
auditory, and textual concepts---represents a promising direction for building more general-purpose memory systems.




\bibliographystyle{named}
\bibliography{ijcai26}

\end{document}